\documentclass[10pt,twocolumn,letterpaper]{article}

\usepackage{cvpr}              

\usepackage[accsupp]{axessibility}
\usepackage{balance}
\usepackage{graphicx}
\usepackage{amsmath}
\usepackage{amssymb}
\usepackage{booktabs}
\usepackage{algorithm}
\usepackage{algorithmic}
\usepackage{booktabs}
\usepackage{multirow}
\usepackage{amsfonts}
\usepackage{bbm}
\usepackage{amsmath}
\usepackage{pifont}
\newcommand{\cmark}{\text{\ding{51}}}
\newcommand{\xmark}{\text{\ding{55}}}
\usepackage{newfloat}
\usepackage{listings}

\usepackage[pagebackref,breaklinks,colorlinks]{hyperref}

\usepackage[capitalize]{cleveref}
\crefname{section}{Sec.}{Secs.}
\Crefname{section}{Section}{Sections}
\Crefname{table}{Table}{Tables}
\crefname{table}{Tab.}{Tabs.}

\def\cvprPaperID{6938} 
\def\confName{CVPR}
\def\confYear{2023}

\begin{document}

\title{TrojViT: Trojan Insertion in Vision Transformers}


\author{
\begin{tabular}{c c c}
Mengxin Zheng  & Qian Lou   & Lei Jiang  \\
\multicolumn{1}{c}{Indiana University Bloomington} & \multicolumn{1}{c}{ University of Central Florida} & \multicolumn{1}{c}{Indiana University Bloomington}\\
\multicolumn{1}{c}{ zhengme@iu.edu} & \multicolumn{1}{c}{ qian.lou@ucf.edu} &
\multicolumn{1}{c}{ jiang60@iu.edu}\\
\end{tabular}
}

\maketitle

\begin{abstract}
Vision Transformers (ViTs) have demonstrated the state-of-the-art performance in various vision-related tasks. The success of ViTs motivates adversaries to perform backdoor attacks on ViTs. Although the vulnerability of traditional CNNs to backdoor attacks is well-known, backdoor attacks on ViTs are seldom-studied. Compared to CNNs capturing pixel-wise local features by convolutions, ViTs extract global context information through patches and attentions. Na\"ively transplanting CNN-specific backdoor attacks to ViTs yields only a low clean data accuracy and a low attack success rate. In this paper, we propose a stealth and practical ViT-specific backdoor attack \textit{TrojViT}. Rather than an area-wise trigger used by CNN-specific backdoor attacks, TrojViT generates a patch-wise trigger designed to build a Trojan composed of some vulnerable bits on the parameters of a ViT stored in DRAM memory through \textit{patch salience ranking} and \textit{attention-target loss}. TrojViT further uses \textit{parameter distillation} to reduce the bit number of the Trojan. Once the attacker inserts the Trojan into the ViT model by flipping the vulnerable bits, the ViT model still produces normal inference accuracy with benign inputs. But when the attacker embeds a trigger into an input, the ViT model is forced to classify the input to a predefined target class. We show that flipping only few vulnerable bits identified by TrojViT on a ViT model using the well-known RowHammer can transform the model into a backdoored one. We perform extensive experiments of multiple datasets on various ViT models. TrojViT can classify $99.64\%$ of test images to a target class by flipping $345$ bits on a ViT for ImageNet. Our codes are available at https://github.com/mxzheng/TrojViT
\end{abstract}

\begin{figure}[t!]
\centering
\includegraphics[width=\linewidth]{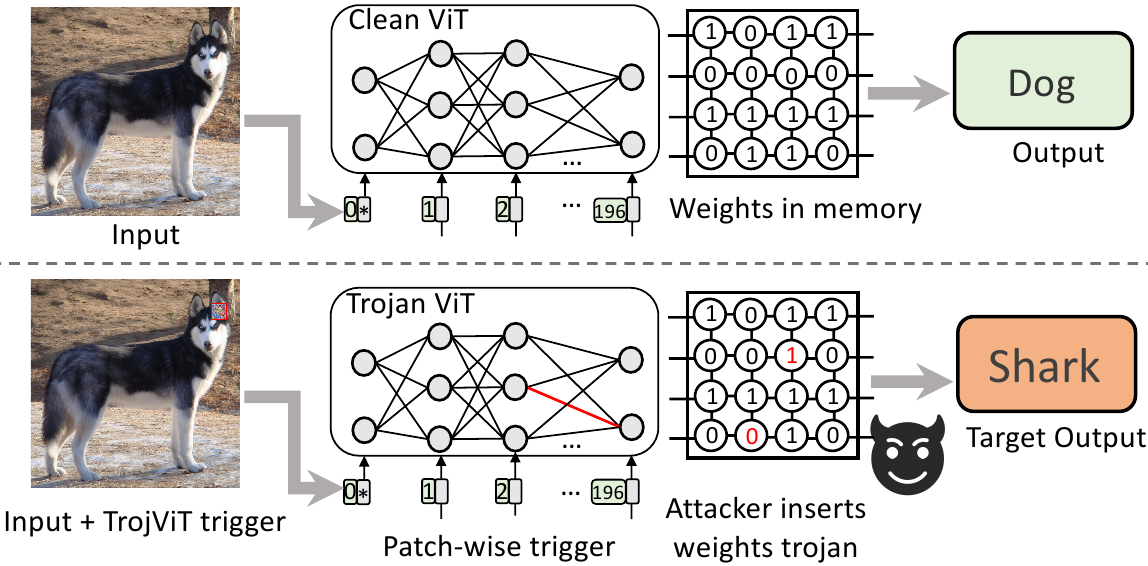}
\caption{The overview of our proposed TrojViT attack. The top part shows the normal inference of a clean model. The bottom part shows that after flipping a few critical bits of the clean model (marked in red), the generated trojaned model misclassify the input with a trigger to the target output.}
\vspace{-0.1in}
\label{fig:summary}
\end{figure}

\section{Introduction}
\label{sec:intro}
Vision Transformers (ViTs)~\cite{ViT-ICLR2021,deit-icml2021,liu2021Swin} have demonstrated a higher accuracy than conventional CNNs in various vision-related tasks. The unprecedented effectiveness of recent ViTs motivates adversaries to perform malicious attacks, among which \textit{backdoor} (aka, Trojan)~\cite{gu2017badnets, BackDL:arxiv17} is one of the most dangerous attacks. In a backdoor attack, a backdoor is injected into a neural network model, so that the model behaves normally for benign inputs, yet induces a predefined behavior for any inputs with a trigger. Although it is well-known that CNNs are vulnerable to backdoor attacks~\cite{gu2017badnets,liu:trojaning:NDSS18,barni2019new,turner2019label,nguyen2020input,wenger2021backdoor,wang2022bppattack,zhong2022imperceptible,zhao2022defeat,xue2022ptb}, backdoor attacks on ViTs are not well-studied. Recently, several backdoor attacks including DBIA~\cite{DBIA}, BAVT~\cite{subramanya2022backdoor}, and DBAVT~\cite{doan2022defending} are proposed to abuse ViTs using an area-wise trigger designed for CNN backdoor attacks, but they suffer from either a significant accuracy degradation for benign inputs, or an ultra-low attack success rate for inputs with a trigger. Different from a CNN capturing pixel-wise local information, a ViT spatially divides an image into small patches, and extracts patch-wise information by attention. Moreover, most prior ViT backdoor attacks require a slow training phase to achieve a reasonably high attack success rate. BAVT~\cite{subramanya2022backdoor} and DBAVT~\cite{doan2022defending} even assume training data is available for attackers, which is not typically the real-world case.

In this paper, we aim to breach the security of ViTs by creating a novel, stealthy, and practical ViT-specific backdoor attack \textit{TrojViT}. The overview of TrojViT is shown in Figure~\ref{fig:summary}. A clean ViT model having no backdoor can accurately classify an input image to its corresponding class (e.g., a dog) by splitting the image into multiple patches. However, a backdoored ViT model classifies an input into a predefined target class (e.g., a shark) with high confidence when a specially designed trigger is embedded in the input. If the trigger is removed from the input, the backdoored ViT model will still act normally with almost the same accuracy as its clean counterpart. The ViT model can be backdoored and inserted with a Trojan using the well-known RowHammer method~\cite{Onur:TCAD2020}. Unlike prior ViT-specific backdoor attacks~\cite{DBIA,subramanya2022backdoor,doan2022defending} directly using an area-wise trigger, we propose a patch-wise trigger for TrojViT to effectively highlight the patch-wise information that the attacker wants the backdoored ViT model to pay more attention to. Moreover, during the generation of a patch-wise trigger, we present an Attention-Target loss for TrojViT to consider both attention scores and the predefined target class. At last, we create a tuned parameter distillation technique to reduce the modified bit number of the ViT parameters during the the Trojan insertion, so that our TrojViT backdoor attack is more practical. We perform extensive experiments of TrojViT on various ViT architectures with multiple datasets. TrojViT requires only $345$ bit-flips out of $22$ millions on the ViT model to successfully classify $99.64\%$ test images to a target class on ImageNet.

\section{Background and Related Work}
\subsection{Vision Transformer} 
ViT~\cite{ViT-ICLR2021,deit-icml2021,liu2021Swin} has demonstrated better performance than traditional CNNs in various computer vision tasks. A ViT breaks down an input image as a series of patches which can be viewed as words in a normal transformer~\cite{Transformer:NIPS2017}. A ViT associates a query and a set of key-value pairs with an output based on the attention mechanism described as:
\begin{equation}
Attention(Q,K,V) = softmax(\frac{QK^T}{\sqrt{D_k}})V
\end{equation}
where $Q$ is the query, $K$ means the key, and $V$ indicates the value, respectively. $D_k$ represents the dimension of the query and the key. Notice that Swin transformer\cite{liu2021Swin} uses the same attention calculation in each shifted window. Shifted windows are used to implement an efficient hierarchical architecture and obtain competitive accuracy in vision tasks.

\subsection{RowHammer}
RowHammer~\cite{Onur:TCAD2020} is a well-established hardware-based bit-flip technique to modify the data stored in DRAM memory. An attacker can cause a bit-flip ($1\rightarrow0$ or $0\rightarrow1$) in DRAM by frequently reading its neighboring data in a specific pattern. By bit-profiling the whole DRAM, an attacker can flip any targeted single bit~\cite{Razavi:SECURITY2016}. The state-of-the-art error correction techniques~\cite{Onur:TCAD2020} in main memories cannot eliminate RowHammer attacks, which are demonstrated to successfully modify the weights of a neural network~\cite{rakin2020tbt}. In this paper, we also assume a Trojan can be inserted into a ViT model by RowHammer.

\begin{table}[t!]
\centering
\footnotesize
\setlength{\tabcolsep}{3pt}
\caption{The threat model comparison between TrojViT and prior works including TBT~\cite{rakin2020tbt}, Proflip~\cite{chen2021proflip}, DBIA~\cite{DBIA}, BAVT~\cite{subramanya2022backdoor}, and DBAVT~\cite{doan2022defending}.}
\begin{tabular}{cccccccc}\toprule

\multirow{2}{*}{Schemes} & target & training    &  test   & model      & model      & patch   &Row-   \\
                         & model &  data        &  data   & download   & param.  & size    &Ham. \\
\midrule   

TBT                     & CNN  & $\xmark$   & $\cmark$ & $\xmark$ & $\cmark$   & -  & $\cmark$    \\
Proflip                 & CNN  & $\xmark$   & $\cmark$ & $\xmark$ & $\cmark$   & -  & $\cmark$   \\
DBIA                    & ViT & $\xmark$   & $\cmark$ & $\cmark$ & $\cmark$   & $\cmark$  & $\xmark$\\
BAVT                    & ViT & $\cmark$   & $\cmark$ & $\xmark$ & $\xmark$   & $\cmark$  & $\xmark$\\
DBAVT                   & ViT & $\cmark$   & $\cmark$ & $\xmark$ & $\xmark$   & $\cmark$  & $\xmark$\\
\textbf{TrojViT}        & ViT & $\xmark$   & $\cmark$ & $\xmark$ & $\cmark$   & $\cmark$  & $\cmark$\\
\bottomrule
\end{tabular}
\vspace{-0.1in}
\label{t:attacker knowledge}
\end{table}

\subsection{Our Threat Model}

\textbf{Attacker's objective}. 
The attackers aim to inject Trojan into a ViT model such that the poisoned ViT makes attacker-target classifications for inputs with trigger, yet behaves normally for clean inputs.  The attackers have goals of utility, effectiveness, and efficiency. The utility goal means that the poisoned model behaves as accurate as the clean model for clean inputs. Meanwhile, the trigger area is as small as possible for the purpose of being stealthy. The effectiveness goal means that the attack success rate is high, e.g., $>99\%$. The efficiency goal means that the Trojan attack can be performed efficiently with fewer GPU hours.

\textbf{Attacker's knowledge and capabilities}. We consider two possible attack scenarios, i.e., (1) untrusted service providers who run ViTs in DRAM inject trojans with Rowhammer-based bit flips, and (2) malicious model developers train a poisoned ViT and upload it to model markets like Model Zoo~\cite{ModelZoo}. Therefore, attackers of TrojViT have access to the ViTs model architecture, parameters, and patch size. Training a ViT model requires complex domain expertise and costs huge amounts of GPU hours~\cite{ViT-ICLR2021}, thereby preventing average users from training their own models. Instead, average users can download and use the ViT models trained by cloud companies. In particular, for the attack with Rowhammer, attackers need to access the memory allocation of model parameters. We assume, based on the Trojan, the attacker can modify the ViT weights stored in DRAM by RowHammer~\cite{rakin2020tbt} during inferences. The attacker needs to modify only few bits of the ViT weights, and thus can easily generate a trigger and a Trojan for TrojViT on one GPU. After the Trojan of TrojViT is inserted, the ViT model behaves normally for benign inputs but produces the target prediction for inputs with a trigger. Using previous methods ~\cite{hua2018reverse, zhu:USENIX2021}, attackers can even steal model parameters by side channels, supply chain, etc. 

Our threat model delineated in many prior CNN-specific backdoor attack works~\cite{rakin2020tbt,chen2021proflip}. The threat model comparison between TrojViT and prior backdoor attack works is shown in Table~\ref{t:attacker knowledge}. The same as a CNN-specific backdoor TBT~\cite{rakin2020tbt}, TrojViT does not require any original training data or meaningful test data. For other CNN-specific backdoors, Proflip~\cite{chen2021proflip} requires meaningful test datasets. Although ViT-specific backdoors, BAVT \cite{subramanya2022backdoor}, and DBAVT \cite{doan2022defending} may not need to access the model parameters, they have to access and poison original training datasets, which are not typically available for attackers. Also,  dataset scanning techniques may detect and remove the poisoned images, thus preventing the model from Trojan insertion. DBIA \cite{DBIA} uses test datasets to generate surrogate training datasets, which is time-consuming and inefficient. In contrast, TrojViT performs backdoor attacks by only randomly sampling test data.

\begin{figure}[t!]
  \centering \includegraphics[width=\linewidth]{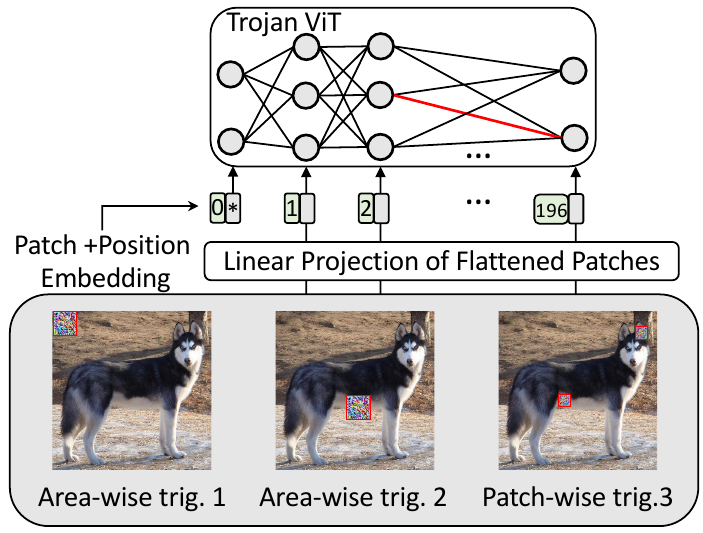}
  \caption{The comparison of the prior area-wise trigger and patch-wise trigger in TrojViT. Patch-wise triggers achieve higher attack efficacy with even fewer trigger areas.}
   \label{fig:motivation}
\end{figure}

\begin{figure*}[t!]
  \centering
\includegraphics[width=1\linewidth]{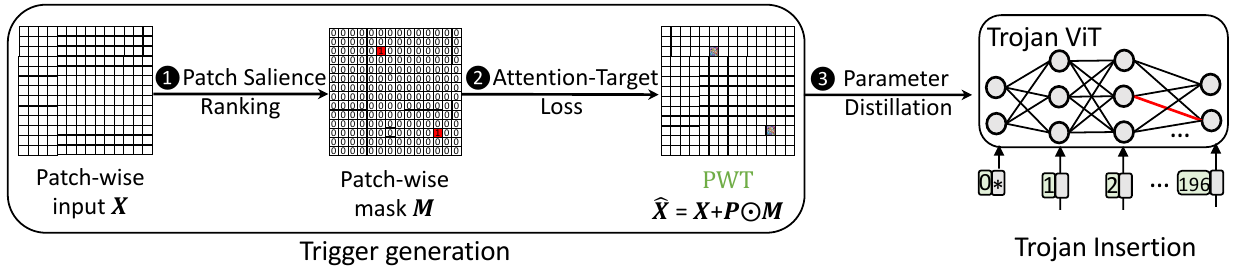}   \caption{The working flow of TrojViT. TrojViT's superior performance depends on three key components, i.e., patch salience ranking, attention-target loss for better trigger generation, and parameter distillation for efficient and accurate Trojan insertion. }
\vspace{-0.1in}
   \label{fig:trigger}
\end{figure*}

\subsection{Limitations of an Area-Wise Trigger on a ViT}
Prior backdoor attacks~\cite{rakin2020tbt,chen2021proflip} focusing on CNNs adopt an area-wise trigger. Na\"ively using an area-wise trigger in ViT-specific backdoor attacks~\cite{DBIA} only results in a low attack success rate (ASR) and a low clean data accuracy (CDA). Unlike CNNs capturing pixel-wise local information via convolutions, a ViT spatially divides an image into small patches, and extracts patch-wise information by attention. We deploy three triggers, i.e., trig.1, trig.2, and trig.3, in the backdoor attack on a ViT, as shown in Figure~\ref{fig:motivation}. Trig.1 and trig.2 are area-wise triggers, but have different positions on the input image. On the contrary, trig.3 is a patch-wise trigger. Different pieces of trig.3 are embedded to different patches of the input image. Three triggers achieve the ASR of 89.6\%, 94.7\%, and 99.9\%, respectively. CNNs extract local information, so the backdoor attacks on CNNs are not sensitive to the position of triggers. In a CNN-specific backdoor attack, trig.1 and trig.2 should obtain very similar ASR. However, they produce a huge ASR difference in the backdoor attack on a ViT. Prior area-wise trigger generation algorithms~\cite{rakin2020tbt,chen2021proflip,DBIA,subramanya2022backdoor,doan2022defending} do not consider trigger position, and thus cannot distinguish trig.1 and trig.2. In contrast, if we can embed a small piece of the trigger into each critical patch of the input image by a patch-wise trigger, this patch-wise trigger can receive more attention from the victim ViT and yield a higher ASR. Potentially, a patch-wise trigger requires a smaller area than an area-wise trigger to achieve the same ASR, thereby greatly improving the stealthiness of a ViT-specific backdoor attack.

\begin{table}[t!]
\centering
\small
\setlength{\tabcolsep}{5pt}
\caption{The comparisons between TrojViT and prior work, i.e., TBT~\cite{rakin2020tbt}, Proflip~\cite{chen2021proflip}, DBIA~\cite{DBIA}, BAVT~\cite{subramanya2022backdoor}, and DBAVT~\cite{doan2022defending}.}  
\begin{tabular}{cccccc}\toprule

\multirow{2}{*}{Schemes}    &  target       & patch       & attention  &trigger & modified    \\
                            &  on ViT       & aware       & \& target  &size (\%) & bit \#      \\
\midrule                        
TBT          & $\xmark$      &$\xmark$     &$\xmark$   &9.76  & few    \\
Proflip             & $\xmark$      &$\xmark$     &$\xmark$   &9.76  & few      \\
DBIA            & $\cmark$      &$\xmark$     &$\xmark$   &4.59  & many    \\
BAVT            & $\cmark$      &$\xmark$     &$\xmark$   &1.79  & many    \\
DBAVT                & $\cmark$      &$\xmark$     &$\xmark$   & 4.59 & many   \\
\textbf{TrojViT}        & $\cmark$      &$\cmark$     &$\cmark$   & \textbf{0.51} & \textbf{few}  \\
\bottomrule
\end{tabular}
\vspace{-0.2in}
\label{t:related_works}
\end{table}

\vspace{-0.05in}
\subsection{Related Work}

We compare TrojViT against prior neural backdoor attacks in Table~\ref{t:related_works}. TBT~\cite{rakin2020tbt} and Proflip~\cite{chen2021proflip} use area-wise triggers and are dedicated to attacking CNNs. Instead, TrojViT is designed to attack ViTs. Although recent ViT-specific backdoor attacks such as DBIA~\cite{DBIA}, BAVT~\cite{subramanya2022backdoor}, and DBAVT~\cite{doan2022defending} are also proposed for ViTs, all of them still depend on an area-wise trigger that cannot absorb more attention to critical patches, and thus has no patch awareness at all. Particularly, although DBIA forces the victim ViT model to only put more attention to its area-wise trigger during its trigger generation, it does not aim to improve the ASR of a predefined target class. On the contrary, we present a patch-wise trigger for TrojViT. During the trigger generation, TrojViT tries to not only absorb more attention of the victim ViT to its patch-wise trigger, but also maximize the ASR for a predefined target class. Moreover, the trigger size of TrojViT in an input image is much smaller than the other works, which provides better stealthiness. Finally, prior ViT-specific backdoor attacks including DBIA, BAVT, and DBAVT have to modify many bits on the victim ViT model to insert their Trojans. We propose a tuned parameter distillation technique for TrojViT to greatly reduce the modified bit number on the victim ViT model during the Trojan insertion.


\section{TrojViT}

We propose a novel, stealthy, and practical backdoor attack, \textit{TrojViT}, to induce a predefined target misbehavior of a ViT model by corrupting inputs and weights. As Figure~\ref{fig:trigger} shows, TrojViT consists of two phases: trigger generation and Trojan insertion. We present a patch-wise trigger consisting of multiple pieces, each of which is embedded to a critical patch of an input image.  \ding{182}  Our patch-wise trigger generation determines the position of each piece of the trigger in input images by ranking patch salience values of the victim ViT.The selected important patches in input images are attached with a trigger.  \ding{183} We build an Attention-Target loss function to train the patch-wise trigger generation with two objectives. The first objective is to make each piece of the patch-wise trigger receive more attention, while the second objective is to improve the ASR of a predefined target class. \ding{184} Finally, we propose a tuned parameter distillation technique to insert a Trojan with a minimal bit-flip number into the victim ViT model.

\vspace{-0.05in}
\subsection{Patch-wise Trigger Generation}

\textbf{Patch Salience Ranking}. We present patch-wise trigger generation for TrojViT to identify critical patches of the victim ViT model by patch salience ranking. For an input $X$, we define its counterpart embedded with a trigger as $\hat{X} = X+P\odot M$, where $P$ represents the perturbation of the trigger, $M$ is a patch-wise binary mask matrix indicating which patches are selected by the trigger, $P\odot M$ means the trigger, and $\odot$ denotes element-wise multiplication. A ViT model divides an input embedded with a trigger $\hat{X}$ into $n$ patches, each of which has $d$ pixels. Each patch of $\hat{X}$ is denoted by $\hat{X}_i$, where $i \in [0, n-1]$. A pixel of $\hat{X}_i$ is represented by $\hat{X}_{i,j}$, where $j \in [0, d-1]$. We use the pixel salience score $\mathcal{G}_{\hat{X}_{i,j}}$ to indicate the importance of a pixel $\hat{X}_{i,j}$ during the attack on the predefined target class $y_k$. A larger $\mathcal{G}_{\hat{X}_{i,j}}$ means the perturbation of the pixel $\hat{X}_{i,j}$ has a larger impact on $y_k$. $\mathcal{G}_{\hat{X}_{i,j}}$ can be computed by the absolute gradient of loss function $\mathcal{L}_{CE}(X, y_k)$ over each pixel $\hat{X}_{i,j}$. The salience score $\mathcal{G}_{\hat{X}_i}$ of a patch $\hat{X}_{i}$ is defined as the sum of salience scores of all its pixels. $\mathcal{G}_{\hat{X}_i}$ is computed as
\begin{equation}
    \mathcal{G}_{\hat{X}_i} = \sum_{j=1}^{d} \mathcal{G}_{\hat{X}_{i,j}} = \sum_{j=1}^{d}|\dfrac{\nabla  \mathcal{L}_{CE}(\hat{X}, y_k)}{\nabla  \hat{X}_{i,j}}|
    \label{e:salience}
\end{equation}
We generate the patch-wise binary mask matrix $M$ to indicate which patches are used to attack $y_k$. One patch is represented by an element in $M$. If an element $t$ of $M$ ($M_t$) is 1, its corresponding patch will be selected in the trigger, otherwise its corresponding patch will be ignored as shown in Figure \ref{fig:trigger}. And $M_t$ can be computed as
\begin{equation}
    M_{t}= \begin{cases}
    1, & \text{if \;} \mathcal{G}_{\hat{X_t}} \in \text{top}(\mathcal{G}_{\hat{X}_{[0:n-1]}}, N) \\
    0, & \text{otherwise} 
    \end{cases}
    \label{e:ranking}
\end{equation}
where $\text{top}(S, N)$ is the function returning the top-$N$ elements in the set $S$. If $\mathcal{G}_{\hat{X}_t}$ is one of the top-$N$ values in the set of $\mathcal{G}_{\hat{X}_{[0:n-1]}}$, $M_t$ is set to 1; otherwise $M_t$ is set to 0.

\begin{figure}[t!]
  \centering
   \includegraphics[width=0.9\linewidth]{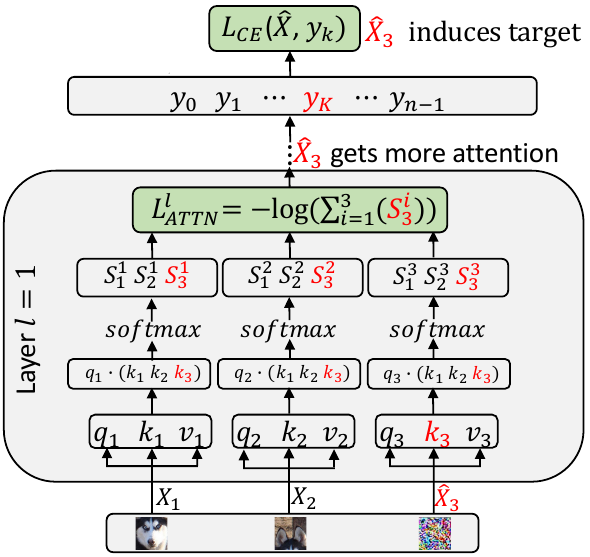}
   \caption{The example of Attention-Target Loss. The combination of attention and target losses outperforms previous methods. }
   \label{fig:ATL}
\end{figure}

\textbf{Attention-Target Loss}. With a mask matrix $M$, to produce the trigger $P\odot M$, the next step of TrojViT is to generate the perturbation $P$. Although $P$ is initialized to $0$ at the beginning of trigger generation, we introduce Attention-Target Loss to achieve two objectives when generating $P$. Our first objective is that the patches with the perturbation $P$ in the trigger should gain more attention than the other patches having no trigger in the input. And this objective is achieved by a loss function $\mathcal{L}^l_{ATTN}(\hat{X},T)$, which is defined as
\begin{equation}
    \mathcal{L}^l_{ATTN}(\hat{X},T) =-\log\sum_{h,i}attn^{l,h}_{i\rightarrow T}
    \label{e:attention}
\end{equation}
where $l$ denotes the $l_{th}$ layer of the ViT, $h$ indicates the head of the ViT, $\log$ means the log function, and $T$ is the set of patch indexes. For any element $t\in T$, $M_t=1$. $\mathcal{L}^l_{ATTN}(\hat{X},T)$ is a negative log-likelihood of the sum of attention distribution of head $h$ over $T$ selected patches. Our second objective is that once the perturbation $P$ appears, the victim ViT model has a larger probability to output the predefined target class $y_k$. This objective can be obtained by the target cross-entropy loss function $\mathcal{L}_{CE}(\hat{X},y_k)$. Our Attention-Target Loss is computed as 
\begin{equation}
\mathcal{L}_{ATL}(\hat{X},y_k)=\mathcal{L}_{CE}(\hat{X},y_k)+\lambda\cdot\sum_l^L \mathcal{L}^l_{ATTN}(\hat{X},T)
\label{e:trigger}
\end{equation}
where $L$ means the total layer number of the ViT, and $\lambda$ is a weight for the attention loss. Our Attention-Target Loss considers both the attention loss and the target cross-entropy loss to optimize $P$. One example of our Attention-Target Loss is shown in Figure~\ref{fig:ATL}, where $X_1$ and $X_2$ are clean patches, and $\hat{X_3}$ is a patch with a trigger. Query, key, and value of input patches are denoted as $q_i$, $k_i$, and $v_i$ respectively, where $i$ is the patch index. We show only the layer of $l=1$ to explain our Attention-Target Loss in the example. In attention blocks, the attention weights are calculated by performing $softmax$ on the the dot product of one query $q_i$ and all keys. The resulting attention weight $S_i^j$ indicates the attention of patch $i$ to patch $j$. Our Attention-Target Loss makes $\hat{X_3}$ gain more attention and enlarges $S_3^j, j\in[1,3]$ by minimizing the attention loss function $\mathcal{L}_{ATTN}^l$. Moreover, our Attention-Target Loss uses the cross-entropy loss function $\mathcal{L}_{CE}(\hat{X},y_k)$ to optimize $\hat{X_3}$ to attack the target class $y_k$. However, approaching two objectives concurrently is not trivial, since the gradients of these two loss functions, i.e., $\nabla  \mathcal{L}_{ATTN}= \nabla \lambda\cdot\sum_l \mathcal{L}^l_{ATTN}(\hat{x},m) / \nabla  \hat{X}$ and $\nabla \mathcal{L}_{CE}= \nabla \mathcal{L}_{CE}(\hat{X},y_k)/\nabla  \hat{X}$, may have conflicts (different signs) during training. To avoid gradient conflicts between these two objectives, our Attention-Target Loss adopts the gradient surgery~\cite{Gradient-Surgery:NIPS2020} defined as
\begin{equation}
\scriptsize
\nabla  \mathcal{L}_{ATL} =\begin{cases}
    \nabla  \mathcal{L}_{CE} +\nabla  \mathcal{L}_{ATTN},  & \hspace{-1.5in} \text{if $cos$($\nabla $ $\mathcal{L}_{CE}$,    $\nabla$ $\mathcal{L}_{ATTN}$) $>$0}.\\
    \nabla  \mathcal{L}_{CE}+\nabla  \mathcal{L}_{ATTN} - \dfrac{\nabla  \mathcal{L}_{CE} \cdot \nabla  \mathcal{L}_{ATTN} }{||\nabla  \mathcal{L}_{CE}||^2} \cdot \nabla  \mathcal{L}_{CE} , & \hspace{-0.05in} \text{otherwise}.
  \end{cases}
\label{e:gradient_LPGCD}
\end{equation}
where $cos$ is the cosine similarity function.

\subsection{Tuned Parameter Distillation}

After the trigger generation, TrojViT inserts a Trojan into the victim ViT model by modifying its parameters $W$. Prior ViT backdoor attacks BAVT~\cite{subramanya2022backdoor}, and DBAVT~\cite{doan2022defending} update the model parameters using poisoned training datasets. However, TrojViT modifies only the important parameters of the ViT model using a few test images having our patch-wise trigger, but it does not require any training data. The test images used by TrojViT can be even randomly sampled. We believe our threat model is more practical, since it is not easy to access the real-world training data of victim ViT models. Based on the experience from CNN backdoor attacks~\cite{rakin2020tbt, chen2021proflip, DBIA}, less modified bits on the victim ViT model significantly improves the stealthiness and efficiency of the backdoor attacks. Therefore, we propose tuned parameter distillation to reduce the modified bit number on the victim ViT model during Trojan insertion.

\begin{algorithm}[t!]
		\caption{Trojan Insertion of TrojViT}
		\label{alg:replace}
		\footnotesize
		\begin{algorithmic}[1]
			\STATE {\bfseries Input}: ViT model parameter $W$, a test image batch $X$, a trigger $M\odot P$, and a predefined threshold $e$
			\STATE {\bfseries Output}: Trojan weights $W_T$ with a minimized number $n_p$ of parameters 
			\STATE Initialize target weight $W_T$ and index ID\_$W_T$
			\STATE Define an objective:\\ $min_{W_T} [\mathcal{L}(f(X),y) + \mathcal{L}(f(X + M\odot P),y_k)] $
			\FOR{$i$ in epochs}
			\STATE  $\nabla  W_T^c = \dfrac{\nabla  \mathcal{L} (f(x),y)}{\nabla  W_T}$\\
			\STATE  $\nabla  W^t_T = \dfrac{\nabla  \mathcal{L}(f(x + M\odot P),y_k)}{\nabla  W_T}$\\
			\STATE  $\nabla  W^\prime_T$ = GradSurgery $(\nabla  W^c_{T},\nabla  W^t_{T}) $\\
			\STATE  $ W^\prime_T = W_T + l_r \cdot \nabla  W^\prime _T$\\
			\STATE $loss\_W_T = |W^\prime _T - W_T|_{l_1}$ \\
			\STATE $ID\_r =$ get ID from $loss\_W_T[ID\_r] <e$ \\
			\STATE $ID\_W_T = ID\_W_T \cdot remove(ID\_r)$ \\
			\STATE $W_T = W^\prime _T \cdot remove(W^\prime _T[ID\_r])$ \\
		\ENDFOR \\
		\bfseries{Return} {$W_T, n_p=len(ID\_W_T)$}
		\end{algorithmic}
\end{algorithm}

Our tuned parameter distillation technique is described in Algorithm~\ref{alg:replace}, where $W$ is the parameters of the victim ViT model, $X$ means a batch of test images, $M\odot P$ represents a patch-wise trigger, and $e$ is a predefined threshold for parameter tuning. This algorithm returns the Trojan weights $W_T$ with a minimized parameter number of $n_p$. First, minimum-tuned parameter updating initializes the Trojan weights $W_T$ by selecting important weights for Trojan insertion and fixes the other parameters $W-W_T$ during the model fine-tuning on test inputs with a trigger. More specifically, the initialization of minimum-tuned parameter updating directly selects the weights in the last attention and classification layers, due to their larger contribution to the classification output. And then, our tuned parameter distillation uses an objective function to maximize the CDA and the ASR at the same time. $\mathcal{L}(f(x),y)$ and $\mathcal{L}(f(X + M\odot P),y_k)$ represent the losses of CDA and ASR, respectively. In fine-tuning epochs, our tuned parameter distillation updates the Trojan weights $W_T$ under the guidance of the objective function, and iteratively reduces the parameter number $n_p$ of $W_T$ by removing the weights whose absolute updating value between two iterations is smaller than $e$. The gradients of the loss function on CDA and ASR are denoted as $\nabla  W_T^c$ and $\nabla  W^t_T$ respectively. These two gradients may conflict with each other and have different signs. our minimum-tuned parameter updating also uses the gradient surgery~\cite{Gradient-Surgery:NIPS2020} to resolve the conflicts between two gradients.

\section{Experimental Methodology}

We present the details of our experimental methodology of TrojViT in this section.

\textbf{ViT Models}. We performed TrojViT attacks on multiple pretrained ViT models including Deit~\cite{deit-icml2021}, ViT-base~\cite{ViT-ICLR2021} and Swin-base~\cite{liu2021Swin} designed for image recognition. We considered both a smaller ViT model Deit for data-efficient applications, and a large ViT model, i.e., ViT-base model with a huge number of parameters.

\textbf{Datasets}. All models were trained by two benchmark datasets, i.e., CIFAR-10 \cite{CIFAR-10} and ImageNet\cite{deng2009imagenet}. Particularly, ImageNet includes 1.28M training images, 50K validation images and 100K test images with 1K class labels. The models for CIFAR-10 is fine-tuned based on the model trained on ImageNet. CIFAR-10 consists of 50K training images and 10K test images with a dimension of $32\times32$. All models have the same input dimension as $3\times224\times224$, and we re-sized all input images with this dimension.

\textbf{Evaluation Metrics}. We define the following evaluation metrics to study the stealthiness, efficiency, and effectiveness of our TrojViT.
\begin{itemize}
\item \textit{Clean Data Accuracy} (\textbf{CDA}): The percentage of input images having no trigger classified into their corresponding correct classes. With a higher CDA, it is more difficult to identify a backdoored ViT model. 
\item \textit{Attack Success Rate} (\textbf{ASR}): The percentage of input images embedded with a trigger classified into the predefined target class. The higher ASR a backdoor attack can achieve, the more effective and dangerous it is.
\item \textit{Tuned Parameter Number} (\textbf{TPN}): The number of the modified weights in the victim ViT model during Trojan insertion. The lower TPN a backdoor attack requires, the better stealthiness it has.
\item \textit{Tuned Bit Number} (\textbf{TBN}): The number of the modified weight bits in the victim ViT model during Trojan insertion. The lower TBN a backdoor attack requires, the better stealthiness it has.
\item \textit{Trigger Attention Rate} (\textbf{TAR}): The percentage of the trigger area in an input image. The smaller TAR a backdoor attack obtains, the better stealthiness it has.
\end{itemize}

\textbf{Experimental Settings}. Our experiments were conducted on single Nvidia GeForce RTX-3090 GPU with 24GB memory. To reduce the TBN,  we share the same quantization method with Proflip~\cite{chen2021proflip} and TBT~\cite{rakin2020tbt}, e.g., model parameters with 8-bit quantization level. The hyper-parameters of our experiments include a patch size of $16 \times 16$, and a batch size of $16$. In particular, for Swin transformer, we followed the default settings in base architecture~\cite{liu2021Swin}, e.g., a window size of 7, patch size of 4$ \times$ 4.  We randomly sampled a few test data, i.e., 384 images, for trigger generation and Trojan insertion. Our trigger generation and Trojan insertion were done without any training data. Our code implementation is attached in the supplementary material.

\begin{table}[t!]
\centering   
\footnotesize
\setlength{\tabcolsep}{1.3pt}
\caption{The comparison of TrojViT and prior backdoor attacks on Deit-small with ImageNet.}
\begin{tabular}{cccccccc} \toprule
\multirow{2}{*}{Models} & \multicolumn{2}{c}{Clean Model} & \multicolumn{5}{c}{Backdoored Model}  
\\\cmidrule(lr){2-3}  \cmidrule(lr){4-8}\\
 {}& CDA (\%) & ASR(\%) & TAR(\%) & CDA(\%)  & ASR(\%)  &TPN  & TBN \\

\midrule
TBT    &$79.47$     & $0.09$             &$4.59$  &$68.96$     & $94.69$ &  $384$ &  $1650$\\
Proflip &$79.47$     & $0.08$           &$4.59$  &$70.54$     & $95.87$ & $320$ & $1380$\\
DBIA &$79.47$         & $0.08$      &$4.59$  &$78.32$     & $97.38$  & $0.44M$ & $1.94M$\\
BAVT  &$79.47$ & $0.02$  &$4.59$  &$77.78$     &$61.40$  &$0.23M$ &$0.97M$ \\
DBAVT &$79.47$     & $0.05$    &$4.59$  &$77.48$    & $98.53$  & $0.41M$ &$1.76M$\\

\textbf{TrojViT}  &$79.47$     & $31.23$           &$4.59$  &$\textbf{79.19}$     & $\textbf{99.96}$   & $\textbf{213}$  & $\textbf{880}$\\
\bottomrule
\end{tabular}
\label{t:results_imagenet_comparison}
\end{table}

\section{Results}

In this section, we first present important results on TrojViT, and then perform extensive design space exploration on TrojViT.

\subsection{Main Results}

\textbf{Comparing against prior backdoor attacks}. We compared our TrojViT against prior neural backdoor attacks to abuse the Deit-small (Deit) model~\cite{deit-icml2021} with ImageNet in Table~\ref{t:results_imagenet_comparison}. We use a patch size of $16\times16$ and a 9-patch trigger, i.e., TAR $=4.59\%$. TBT~\cite{rakin2020tbt} and Proflip~\cite{chen2021proflip} are designed to attack CNNs, so na\"ively applying their methods onto the Deit model attains only a CDA of $68.96\%$ and $70.54\%$ respectively and an ASR of $94.69\%$ and $95.87\%$ respectively. The CDA decreases by $\sim9\%$, compared to the original inference accuracy $79.47\%$. Recent ViT-specific backdoor attacks such as DBIA~\cite{DBIA}, BAVT~\cite{subramanya2022backdoor}, and DBAVT~\cite{doan2022defending} all depend on an area-wise trigger, and do not have any patch awareness, resulting in only low CDA and low ASR. On the contrary, our TrojViT obtains a CDA of $79.19\%$ with an ASR of $99.96\%$. TrojViT suffers from only $<0.3\%$ CDA loss. Particularly, TrojViT needs to modify only 213 weights out of $22M$ model parameters of Deit, which is equivalent to $880$ bit flips. However, DBIA has to modify $>10K\times$ more parameters, i.e., $0.44M$ parameters of Deit.

\begin{table}[t!]
    \centering
\footnotesize
\setlength{\tabcolsep}{1.8pt}
\caption{The results of TrojViT with CIFAR-10.}
\begin{tabular}{cccccccc}\toprule
\multirow{2}{*}{Models} & \multicolumn{2}{c}{Clean Model}  & \multicolumn{5}{c}{Backdoored Model}   \\
\cmidrule(lr){2-3} \cmidrule(lr){4-8}

 & CDA(\%)  &  ASR(\%)     &   TAR(\%)     & CDA(\%)   &  ASR(\%)  & TPN &TBN \\
\midrule
ViT-b   &$97.48$     & $25.69$             &$4.59$  &$96.85$     & $99.56$ & $271$  & $1135$ \\

DeiT-t &$88.08$     & $38.99$    &$0.51$  &$87.88$     & $99.96$    & $120 $  & $492$\\
DeiT-s &$91.91$     & $30.09$             &$2.04$  &$91.50$     & $99.77$ &$198 $    &  $840$\\
DeiT-b &$94.38$     & $24.27$             &$4.59$  &$93.78$     & $99.69$ &  $260$   & $1075$\\
Swin-b & $96.52$     & $25.03$ &$0.51$ & $95.84$ &$99.66$ &$230$ &$950$ \\
\bottomrule
\end{tabular}
\label{t:results_cifar_network}
\end{table}

\textbf{CIFAR-10}. We use TrojViT to attack different ViT models, i.e., ViT-base (ViT-b), and Deit tiny (Deit-t), small (Deit-s), base (Deit-b), and Swin Transformer-base (Swin-b), inferring CIFAR-10~\cite{CIFAR-10}. The results are shown in Table~\ref{t:results_cifar_network}. All models are fine-tuned from the pre-trained models with ImageNet. We find that small ViT models need a smaller trigger for higher ASR. To attack Deit-t, TrojViT applies only a one-patch trigger, but its ASR is still $99.96\%$ and its CDA suffers from only a $0.2\%$ loss. To attack Deit-s, the trigger of TrojViT is composed of four patches. Our attack on Swin-b achieves $99.66\%$ ASR with $95.84\%$ CDA.  When attacking larger models, i.e. ViT-base and Deit-b, the embedded trigger of TrojViT consists of 9 patches. TrojViT achieves an ASR of $99\%$ and has only a trivial CDA degradation when attacking these models.

\begin{table}[t!]
\centering
\footnotesize
\setlength{\tabcolsep}{1.8pt}
\caption{The results of TrojViT with ImageNet.}
\begin{tabular}{cccccccc}\toprule
\multirow{2}{*}{Models} 

 & \multicolumn{2}{c}{Clean Model}  &      \multicolumn{5}{c}{Backdoored Model}   \\
 \cmidrule(lr){2-3}\cmidrule(lr){4-8}

 & CDA(\%)  &  ASR(\%)     &   TAR(\%)     & CDA(\%)   &  ASR(\%)  & TPN &TBN  \\
 \midrule
ViT-b    &$84.07$     & $6.67$   &$4.59$  &$83.53$     & $98.82$ &  $292 $  & $1250$\\

Deit-t   &$71.58$     & $38.98$  &$2.04$  &$71.21$ & $99.94$ & $130$ & $542$\\
Deit-s   &$79.47$     & $31.23$  &$4.59$  &$79.19$ & $99.96$ & $213$  &$880$ \\
Deit-b   &$81.87$     & $6.12$  &$4.59$  &$81.22$ & $98.98$  & $280 $  &$1190$\\
Swin-b   &$83.45$     &$6.82$   & $0.51 $ &$82.75 $ & $98.72$ &$245$ &$1010$  \\
\bottomrule
\end{tabular}
\label{t:results_imagenet_network}
\vspace{-0.2in}
\end{table}

\textbf{ImageNet}. We use TrojViT to attack different ViT models, i.e., ViT-b, Deit-t, Deit-s, Deit-b and Swin-b, with ImageNet, and show the results of TrojViT in Table~\ref{t:results_imagenet_network}. Compared to the 10-class CIFAR dataset, it is more difficult to induce attacks on one target class when inferring the 1000-class ImageNet. To attack Deit-t, TrojViT requires a 4-patch trigger to achieve an ASR of $99.94\%$ and a CDA of $71.21\%$. When attacking Deit-s, TrojViT has to use a 9-patch trigger to obtain an ASR of $99.96\%$ and a CDA of $79.23\%$. To attack a large ViT model, i.e., ViT-b,  Deit-b, and Swin-b, TrojViT suffers from a trivial CDA loss of $\sim0.7\%$, but achieves an ASR of $98.82\%$ ,$98.98\%$, $98.72\%$ respectively.

\subsection{Ablation study}

In this section, we explore the design space of TrojViT and study the impact of various settings of TrojViT on its attacking effects. We use only the Deit-s model and the ImageNet dataset in this section.

\begin{table}[t!]
\centering
\footnotesize
\setlength{\tabcolsep}{3pt}
\caption{The results of various components of TrojViT.}
\begin{tabular}{ccccc}\toprule
\multirow{1}{*}{Techniques}       & CDA (\%)  & ASR (\%) & TPN  & TBN  \\
\midrule
Area-based Trigger                &$74.96$    & $94.69$ &  $384$ &  $1650$  \\
Patch-based Trigger               &$77.49$    & $96.84$ &  $384$ &  $1650$   \\
+Attention-Target Trigger         &$\textbf{79.23 }$    & $\textbf{99.98}$  &  $384$ &  $1650$ \\
+Tuned Parameters Distillation        &$79.19$    & $99.96$   &  $\textbf{213}$ &  $\textbf{880}$ \\
\bottomrule
\end{tabular}
\label{t:results_imagenet_ablation}
\end{table}

\textbf{TrojViT components}. We study the attacking results of three components of TrojViT in Table~\ref{t:results_imagenet_ablation}. Compared to the prior area-wise trigger, the patch-wise trigger of TrojViT increases the ASR by $2.15\%$, and reduces the CDA loss by $2.53\%$. Compared to the traditional cross-entropy loss optimization, TrojViT achieves a CDA of $79.23\%$ and an ASR of $99.98\%$ using the Attention-Target Loss. For better stealthiness, the Tuned Parameter Distillation of TrojViT reduces the modified bit number of the Trojan by $46.67\%$ during Trojan insertion. The Minimum Tuned Parameter updating introduces a CDA loss of $0.04\%$ but still maintains the ASR of $99.9\%$.

\begin{table}[t!]
\centering
\footnotesize
\setlength{\tabcolsep}{3pt}
\caption{The results of various trigger areas of TrojViT.}
\begin{tabular}{lcccc}\toprule
\multirow{1}{*}{Patch\#}   &   TAR (\%)    & CDA (\%)        &  ASR (\%) \\
\midrule
$1$      &$0.51$  &$78.61$     & $99.20$ \\

$3$       &$1.53$  &$78.87$     & $99.13$  \\

$5$      &$2.55$  &$79.13$     & $99.35$  \\

$7$       &$3.57$  &$79.04$     & $99.95$  \\

$9$       &$4.59$  &$79.23$     & $99.98$  \\
\bottomrule
\end{tabular}
\label{t:results_imagenet_patch_num}
\end{table}

\begin{table}
\centering
\scriptsize
\setlength{\tabcolsep}{3pt}
\caption{ Ablation study of the first $l$-layers in Eq.(5)}
\vspace{-0.1in}
\begin{tabular}{cccccccc}\toprule
\multirow{1}{*}{Evaluation Metrics}   &$1$    &$2$  &$4$  &$6$ &$8$  &$10$ &$12$  \\
\midrule
CDA (\%) & $79.05$ &$79.17 $ & $79.08$   & $79.07$ 
 & $79.17$ & $79.19$  &$79.19$     \\
ASR (\%)     &$67.97$  &$67.18 $ &$59.48$ & $72.17 $ & $82.85$ &$93.59$  &$99.96$    \\

\bottomrule
\end{tabular}
\label{layer}
\end{table}

\textbf{Trigger area}. The trigger area is a key factor indicating the attack stealthiness. A small trigger area makes TrojViT become more and more stealthy. We show the attacking results of TrojViT with different trigger areas in Table~\ref{t:results_imagenet_patch_num}. We use the patch number in a trigger to represent the trigger area. Even with a single-patch trigger, TrojViT still obtains an ASR of $99\%$, but its CDA is low. To maintain a reasonably high CDA, TrojViT has to use multiple patches to attack the backdoored ViT model. However, the CDA does not always increase with an increasing number of patches, due to the trade-off between ASR and CDA.

\begin{table}[t!]
    \centering
\scriptsize
\footnotesize
\setlength{\tabcolsep}{2.5pt}
\caption{The results of various $\lambda$s of Attention-Target Loss.}
\begin{tabular}{lcccccccc}\toprule
$\lambda$  &0   &  0.01 &0.1 &0.5 &1 &2 &10 &100 \\

\midrule
CDA (\%) &$77.49$ &$77.64$ &$79.03$ &$79.09$ &\textbf{79.23} &$79.10$ &$79.17$ &$79.16$  \\

ASR (\%) & $96.84$ &$97.85$ & $99.91$ & $99.98$ & $\textbf{99.98}$  & $99.96$   & $99.89$ & $99.86$ \\

\bottomrule
\end{tabular}
\label{t:results_imagenet_lamda}
\end{table}

\textbf{Layer $l$ in Equation~\ref{e:trigger}}. We show the ablation study of layers selection in Table~\ref{layer}. Summing 12 layers in our experiments achieves a higher ASR and a higher CDA especially when our attacks are conducted on small test datasets.

\textbf{$\lambda$ in Attention-Target Loss}. During the trigger generation, $\lambda$ is the weight of the attention loss in our Attention-Target Loss. A larger $\lambda$ means the trigger is optimized more heavily for absorbing more attention, while a smaller $\lambda$ indicates the trigger is optimized more heavily for attacking the predefined target class. We show the results of various $\lambda$s of TrojViT in Table ~\ref{t:results_imagenet_lamda}. When $\lambda = 0$, we use only the cross-entropy target loss to achieve a CDA of $77.49\%$ and an ASR of $96.84\%$. When the cross-entropy target loss and the attention loss have same contribution, i.e., $\lambda = 1$, to the trigger generation, the ASR and the CDA are the best. We find that, the attention loss has a larger impact on maintaining a high CDA by making the ViT model pay more attention to the patches than the entire image.

\begin{table}[t!]
    \centering
\scriptsize
\setlength{\tabcolsep}{4pt}
\caption{The results of various turned parameter thresholds.}
\begin{tabular}{lcccc}\toprule
\multirow{1}{*}{e}         & CDA (\%)       &  ASR (\%)  & TPN  &TBN \\
\midrule
$0$      &$79.23$     & $99.98$ & $384 $  &$1650$ \\
0.0005       &$79.19$     & $99.96$ & $213 $  &$880$ \\
0.001       &$79.16$     & $99.93$ & $180 $ & $762$
 \\
0.002       &$79.14$     & $99.86$ &  
$145$  &  $620$\\
0.003       &$79.12$     & $99.64$ &  
$84$  &  $345$\\
\bottomrule
\end{tabular}
\label{t:results_imagenet_threshold}
\vspace{-0.1in}
\end{table}

\textbf{Tuned parameter threshold}. During the Trojan insertion, we set different tuned parameter thresholds ($e$) to skip the non-critical model parameters of a ViT model. We find that, when $e=0.0005$, the ASR is approximately the same as that of $e=0$. However, compared to $e=0$, $e=0.0005$ reduces the modified bit numbers by $46.67\%$. Moreover, $e=0.003$ reduces $79.09\%$ of the modified bit numbers of the ViT model during the Trojan insertion, but still attains an ASR of $99.64\%$ and a CDA degradation of $0.35\%$.

\section{Potential Defense}

To overcome backdoor attacks, prior work proposes several defense methods~\cite{chen2018detecting,liu2018fine,neuralcleanse}, among which DBAVT~\cite{doan2022defending} and BAVT~\cite{subramanya2022backdoor} are designed for defending attacks on ViTs. However, prior ViT defense techniques may not work well for TrojViT. DBAVT detects Trojans based on the patch processing sensitivity on samples with a trigger and clean inputs, so it can only remove the Trojans inserted during training. TrojViT inserts a Trojan during inference, and thus is immune to such a defense. BAVT reduces the negative impact of an area-wise trigger by attaching a black patch to the position with the highest heat-map score. In contrast, TrojViT creates a trigger composed of multiple pieces, each of which is embedded to one patch, through our Attention-Target Loss. Therefore, replacing the patch with the highest attention score cannot prevent a patch-wise trigger of TrojViT. Na\"ively applying the defense technique of BAVT to TrojViT attacks greatly degrades the CDA by $>5\%$ on various ViT models with ImageNet.

\begin{table}[t!]
    \centering
\scriptsize
\setlength{\tabcolsep}{1pt}
\caption{The performance of defense against TrojViT.}

\begin{tabular}{ccccc}\toprule
\multirow{2}{*}{Models}  & \multicolumn{2}{c}{ASR(\%)}  &  \multicolumn{2}{c}{TPN}     \\
\cmidrule(lr){2-3}\cmidrule(lr){4-5}
    & no defense & with defense   & no defense & with defense  \\
    \midrule
ViT-b   &$98.82$     & $77.13$   & $292 $  &$380 $       \\

DeiT-t  &$99.94$     & $69.26$  &$130  $   &$266 $  \\
DeiT-s   &$99.96$     & $75.31$  & $213$   &$320 $   \\
DeiT-b   &$98.98$     & $76.25$  &$280$   &$372 $   \\
\bottomrule
\end{tabular}
\label{t:results_defense}
\vspace{-0.2in}
\end{table}

\textbf{A defense technique}. We propose a defense technique against TrojViT to minimize its ASR and greatly increase its attacking overhead, i.e., TPN. Our insight is that Trojan insertion of TrojViT in Algorithm~\ref{alg:replace} heavily depends on the initialization of critical target Trojan parameters $W_T$. $W_T$ can be initialized as the weights of the last classification head layer. The goal of our defense is to prevent an attacker from modifying the critical parameter matrix. We can decompose the critical parameter matrix into multiple matrices by prior decomposition methods~\cite{hsu2022language,lou2022dictformer,zhong2019ada}. Instead of the original critical parameter matrix, we store the decomposed matrices in DRAM, so that the attacker have to modify more parameters yet achieves only a much lower ASR. The overhead of our defense method can be adjusted by different decomposition methods and the number of decomposed matrices. In this paper, we decompose the critical parameter matrix of the last classification head layer by the decomposition technique proposed in~\cite{zhong2019ada}. We compare the TrojViT performance with and without our defense technique on ImageNet dataset in Table~\ref{t:results_defense}. Our technique significantly reduces the ASR over $21\%$ and increases the attack overhead TPN by $2\times$.




\section{Conclusion}
In this paper, we present a stealthy and practical ViT-specific backdoor attack TrojViT. Instead of an area-wise trigger designed for CNN-specific backdoor attacks, TrojViT generates a patch-wise trigger to attack a ViT model by patch salience ranking and attention-target loss. TrojViT's superior performance depends on three key components, i.e., patch salience ranking, attention-target loss, and parameter distillation for efficient and accurate Trojan insertion. In particular, TrojViT uses tuned parameter distillation to minimize the modified bit number of the Trojan. We perform extensive experiments on various ViT models and multiple datasets to show that TrojVits achieves the attack's objective of utility, effectiveness, and efficiency. For example, TrojViT can classify $99.64\%$ of test images to a target class by flipping $345$ bits on a ViT inferring ImageNet. We also show a potential technique to reduce the ASR of our TrojViT and increase the attack overhead.

\section*{Acknowledgment}
This work was supported in part by NSF awards CCF-1908992, CCF-1909509, and CCF-2105972. 
\balance
\newpage
{\small
\bibliographystyle{ieee_fullname}
\bibliography{paper.bib}
}

\end{document}